\documentclass[10pt,twocolumn,letterpaper]{article}

\usepackage{iccv}
\usepackage{times}
\usepackage{epsfig}
\usepackage{graphicx}
\usepackage{amsmath}
\usepackage{amssymb}
\usepackage{subcaption}
\usepackage{lipsum}
\usepackage{mathptmx}
\usepackage[square,numbers]{natbib}
\usepackage[pagebackref=true,breaklinks=true,letterpaper=true,colorlinks,bookmarks=false]{hyperref}

\iccvfinalcopy 

\def\iccvPaperID{0034} 

\ificcvfinal\fi

\usepackage{todo}
\usepackage{xcolor}

\graphicspath{{images/}}

\begin{document}

\title{Fourier-CPPNs for Image Synthesis}

\author{Mattie Tesfaldet, Xavier Snelgrove, David Vazquez\\
Element AI\\
Montr\'eal, Canada\\
{\tt\small {mattie,xavier.snelgrove,dvazquez}@elementai.com}}

\maketitle

\ificcvfinal\thispagestyle{empty}\fi

\begin{abstract}
    Compositional Pattern Producing Networks (CPPNs) are differentiable networks that independently map $(x,y)$ pixel coordinates to $(r,g,b)$ colour values. Recently, CPPNs have been used for creating interesting imagery for creative purposes, \eg neural art. However their architecture biases generated images to be overly smooth, lacking high-frequency detail. In this work, we extend CPPNs to explicitly model the frequency information for each pixel output, capturing frequencies beyond the DC component. We show that our Fourier-CPPNs (F-CPPNs) provide improved visual detail for image synthesis.
\end{abstract}

\section{Introduction}
\label{sec:introduction}

The fields of computer graphics and computer vision have a long history of introducing new computational approaches for the creation of images.
Recently, exciting new deep learning approaches for synthesizing images have come about, such as Generative Adversarial Networks (GANs)~\cite{goodfellow2014generative} for generating realistic faces~\cite{karras2018progressive, karras2018style}, Convolutional Networks (ConvNets) for image style transfer~\cite{gatys2016image, ulyanov2016texture, johnson2016perceptual},
and Compositional Pattern Producing Networks (CPPNs)~\cite{stanley2007compositional, karpathy2014image, ha2016generating, metz2017compositional, mordvintsev2018differentiable, snelgrove2018interactive} for creating aesthetically interesting high-resolution images for creative purposes. The proposed research extends CPPNs by explicitly modelling frequency information. Our experiments show that our Fourier-CPPN (F-CPPN) produces images with improved visual detail while maintaining the advantages of CPPNs.

\begin{figure}[t]
    \centering
    \epsfig{file=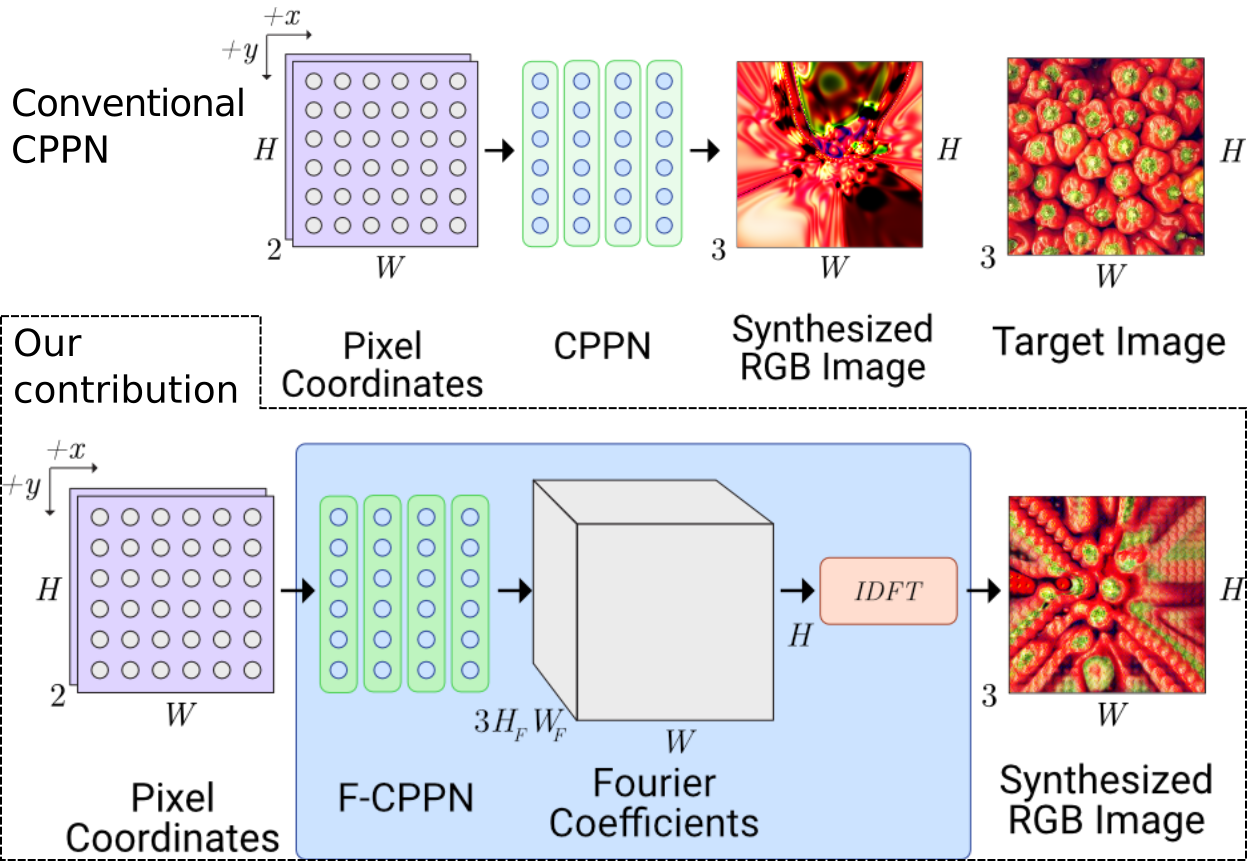, width = \columnwidth}\\
    \caption{Fourier-CPPNs (F-CPPNs) for image synthesis. (top-left) CPPNs are differentiable networks that map $(x,y)$ pixel coordinates to $(r,g,b)$ colour values via linear and non-linear transformations. (bottom) We propose F-CPPNs, an extension of CPPNs which explicitly model the frequency information for each pixel output, capturing frequencies beyond what can be captured by CPPNs. This allows for outputs with increased visual detail.}
    \vspace{-3mm}
    \label{fig:teaser}
\end{figure}

\subsection{Compositional Pattern Producing Networks (CPPNs)}

CPPNs are differentiable networks that independently map $(x,y)$ pixel coordinates to $(r,g,b)$ colour values via the composition and combination of various simple activation functions (Fig.~\ref{fig:teaser} top), \eg linear, exponential, periodic, etc.
Originally, CPPNs were designed for analysing the properties
of natural developmental encodings~\cite{stanley2007compositional} and were optimized using a neural-evolutionary approach that augmented the CPPN's weights \textit{and} topology. Instead of using the same activation function at each layer, the evolutionary process would find the optimal activation from a list of possible activation functions and ``grow'' the network, increasing its complexity. Recently however, CPPNs have started to be used for creating interesting imagery for creative purposes~\cite{ha2016generating,metz2017compositional,mordvintsev2018differentiable,snelgrove2018interactive}, \eg, neural art. These recent CPPN implementations have avoided using neural-evolutionary approaches in determining the network's architecture and have opted for using a fixed architecture instead, whose weights are optimized via gradient descent.

CPPNs have several useful properties. The image they parameterize can be generated at arbitrary resolutions and at arbitrary crops. They can also easily integrate into computer graphics pipelines \cite{snelgrove2018interactive}.
However, the locality of pixel coordinates and choice of smooth activation functions constrains the resulting image such that colour tends to smoothly vary across neighbouring pixels. As a consequence of this inductive bias, CPPNs create images that appear overly smooth, lacking high-frequency detail that can otherwise add realism to the synthesized image. Up until now, CPPNs have not been designed to incorporate frequency information beyond the DC (zero-frequency) component.
We propose an extension to CPPNs, based on Fourier analysis, where each pixel's colour value is represented as a linear combination of complex-valued sinusoids.

\subsection{Fourier synthesis}
Here we briefly review the relevant theory and mathematics before introducing F-CPPNs. The two-dimensional (2D) discrete Fourier transform (DFT) allows us to represent an image, $I$, as a linear combination of complex-valued sinusoidal basis images with varying frequency. The mixing coefficients of these images are given by the complex-valued  $F[\omega_x, \omega_y]$, which represent the magnitude and phase of the sinusoid with spatial frequency $\omega_x$ and $\omega_y$.
The inverse 2D DFT (IDFT) allows us to synthesize $I$ from its Fourier coefficients, $F$. It is defined, per colour channel, as follows,
\begin{equation}
    \small
    I_c(x,y) = \frac{1}{\sqrt{WH}} \
    \sum_{\omega_x=0}^{W-1} \sum_{\omega_y=0}^{H-1} \
    F_c[\omega_x, \omega_y] e^{i2\pi(\omega_x x/W + \omega_y y/H)}\ ,
\label{eq:fourier_synthesis}
\end{equation}
where $I_c(x,y)$ is the intensity at pixel $(x,y)$ for a particular colour channel $c$, $F_c[\omega_x, \omega_y]$ is the Fourier coefficient for the given spatial frequencies
$\omega_x$ and $\omega_y$, and $W$ and $H$ are the width and height of the image in pixels, respectively. Note that the number of frequencies (and coefficients) corresponds to the number of pixels in the input image. $F$ and $I$ above are complex-valued functions. Since we seek to generate a real-valued image, in this work we simply take the real part of $I$.

\section{Fourier CPPNs}
\label{sec:technical_approach}

We propose the \textit{Fourier}-CPPN (F-CPPN), an alternate parameterization to CPPNs where each pixel's $(r,g,b)$ colour value is obtained from an IDFT on a set of learned Fourier coefficients (see Fig.~\ref{fig:teaser} bottom),
\begin{equation}
    \small
    (x, y) \xrightarrow{F-CPPN} \left(F_r[\omega_x, \omega_y], F_g[\omega_x, \omega_y], F_b[\omega_x, \omega_y)\right] \xrightarrow{IDFT} (r,g,b)\ .
    \label{eq:fourier_cppn}
\end{equation}
Consider a $H \times W \times 2$ grid of pixel coordinates as inputs to a F-CPPN. Recall that a $H \times W$ image can be defined by a single set of Fourier coefficients of the same dimensions. We design our F-CPPN to output a smaller number
of coefficients with dimensions $H_F \times W_F$. However, we allow them to vary at each pixel coordinate. This localized and spatially-varying Fourier parameterization of the image is defined as follows,
\begin{multline}
    \small
    I_c(x,y) = \frac{1}{\sqrt{W_F H_F}} \\
	\sum_{\omega_x=0}^{W_F-1} \sum_{\omega_y=0}^{H_F-1}\
		F_{xyc}[\omega_x, \omega_y] e^{i2\pi(\omega_x x/W_F + \omega_y y/H_F)}\ ,
    \label{eq:localized_fourier}
\end{multline}
where $F_{xyc}[\omega_x, \omega_y]$ represents the localized frequency information for spatial frequencies $\omega_x$ and $\omega_y$ at pixel coordinate $(x, y)$ for colour channel $c$. The final $(r,g,b)$ colour value is constructed as $I(x,y) = (I_r(x,y), I_g(x,y), I_b(x,y))$, from which the final $H \times W \times 3$ image $I$ is obtained.

This image representation is overparameterized. Instead of representing the image with $H \times W$ Fourier
coefficients, it is now represented with $H \times W \times H_F \times W_F$ localized Fourier coefficients.
However, this parameterization has some useful properties. First, a region of an image containing a periodic texture with period $W_F$ and $H_F$ in $x$ and $y$, respectively, can be represented with a constant set of localized Fourier coefficients at every pixel location in that region. Second, a region of transition from one periodic texture to another can be represented as an interpolation from one set of localized Fourier coefficients to another.
By combining the inductive bias of a CPPN, which tends towards constant or smoothly varying outputs, with the
property of the localized IDFT, whereby regions with constant coefficients become regions of constant
periodic texture, we arrive at our contribution, the F-CPPN. Our F-CPPN is able to explicitly model frequency information beyond the DC component.
We can consider CPPNs as a special case of F-CPPNs, where $W_F=W_H=1$,
in other words, a F-CPPN that only captures the DC component.

\paragraph{Architecture design}
\label{sec:architecture_design}
Our F-CPPN architecture builds on the CPPN implementation from \citet{mordvintsev2018differentiable}. Their network consists of eight $1 \times 1$ convolutional layers, each with 24 filters and each followed by an activation function $\phi(a) = \left(\arctan(a)/0.67,\ \arctan^2(a)/0.67\right)$, where $a$ is the output from a convolution and $(\cdot,\ \cdot)$ is a channel-wise concatenation. Note that CPPNs can be implemented as ConvNets strictly using $1 \times 1$ convolutions. This implies that no information is shared between neighbouring pixels. Our F-CPPN differs from their CPPN in that the final layer does not directly output $(r,g,b)$ colour values but instead outputs $\left(F_{xyr}[\omega_x, \omega_y], F_{xyg}[\omega_x, \omega_y], F_{xyb}[\omega_x, \omega_y]\right)$ Fourier coefficients, which are then fed to an IDFT to produce $(r,g,b)$ colour values.

\section{Experiments}
\begin{figure*}[t]
    \centering
    \begin{subfigure}[t]{0.33\textwidth}
        \centering
        \epsfig{file=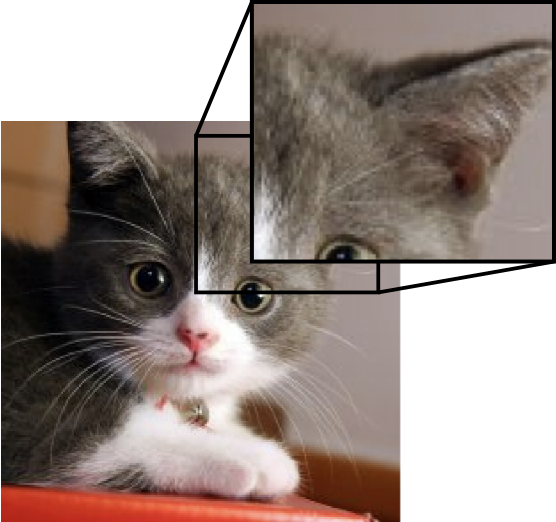, height = 4.5cm}\\
        \label{fig:image_reconstruction_a}
    \end{subfigure}%
    ~
    \begin{subfigure}[t]{0.33\textwidth}
        \centering
        \epsfig{file=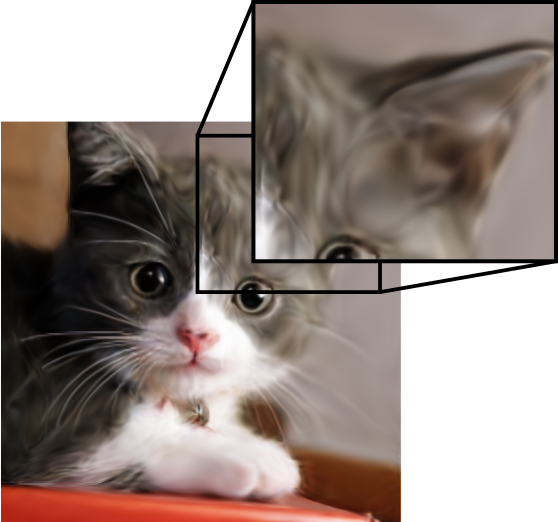, height = 4.5cm}\\
        \label{fig:image_reconstruction_b}
    \end{subfigure}%
    ~
    \begin{subfigure}[t]{0.33\textwidth}
        \centering
        \epsfig{file=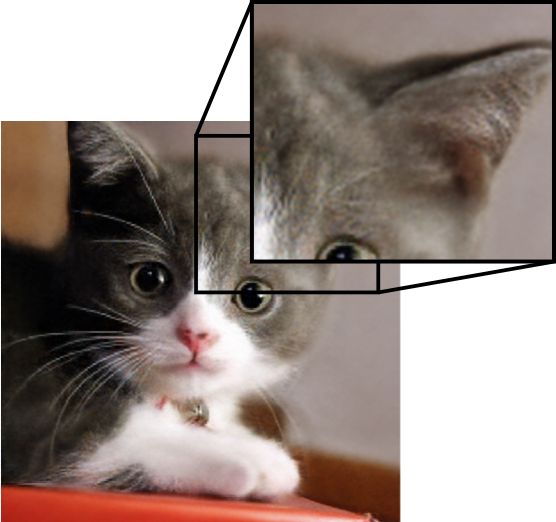, height = 4.5cm}\\
        \label{fig:image_reconstruction_c}
    \end{subfigure}%
    \caption{Image reconstruction via Compositional Pattern Producing Networks (CPPNs) and Fourier-CPPNs (F-CPPNs). (left) Target image. (middle) \citet{mordvintsev2018differentiable}'s CPPN output. (right) Our F-CPPN's output. Notice in the zoomed-in sections that our F-CPPN is able to better reconstruct the textural detail of the cat's fur.}
    \vspace{-3mm}
    \label{fig:image_reconstruction}
\end{figure*}

\label{sec:experiments}

Our goal is to improve the visual detail of images synthesized by CPPNs by explicitly modelling frequency information, arriving at F-CPPNs. We qualitatively evaluate the F-CPPN approach of extending a CPPN's frequency representation beyond the DC component through an ablation study on two image synthesis tasks, image reconstruction and texture synthesis. For each task, we compare the outputs from our F-CPPN and \citet{mordvintsev2018differentiable}'s CPPN (the baseline). Additional results can be found in the appendix.

\paragraph{Training}
The objectives used for optimizing the weights of the F-CPPN and the baseline are described in the following sections. We used L-BFGS \cite{byrd1995limited} for optimization. Results were generated using an NIVIDA Tesla P100 GPU and optimization took about an hour for generating a $224 \times 224$ image. The input pixel coordinates were set to range between $[\sqrt{3}, -\sqrt{3}]$ with $(0, 0)$ in the centre. The weights for each layer were initialized randomly with zero mean and a variance equal to $\sqrt{1/C}$, where $C$ is equal to the number of input activations. Biases were initialized to zero. The number of spatial frequencies, $H_F \times W_F$, was set to $10 \times 10$.

\subsection{Image reconstruction}

We show that F-CPPNs have the capacity to synthesize images with greater detail than the baseline with the straightforward task of image reconstruction. Both a F-CPPN and a CPPN are tasked with reconstructing a given image, $I$, to produce an output, $\hat{I}$. To optimize the weights of both networks, we use the content loss from \citet{gatys2016image},
\begin{equation}
    \mathcal{L}_{content} = \frac{1}{L}\sum_l{||\phi_l(I) - \phi_l(\hat{I})||_2^2}\ ,
    \label{eq:content_loss}
\end{equation}
where $\phi_l(\cdot)$ are the activations of the $l$-th layer of VGG-19~\cite{simonyan2014very} when processing input $(\cdot)$, $L$ is number of layers used, and $||\cdot||_2$ is the L2 norm. In short, the content loss is computed as the mean squared error (MSE) between feature representations of $I$ and $\hat{I}$. The activations used were from layers \emph{conv1\_1}, \emph{pool1}, \emph{pool2}, \emph{pool3}, and \emph{pool4} of VGG-19. As shown in Fig.~\ref{fig:image_reconstruction}, the level of detail captured by the F-CPPN is generally greater than the CPPN's.

\subsection{Texture synthesis}
\begin{figure*}[t]
    \centering
    \begin{subfigure}[t]{0.33\textwidth}
        \centering
        \epsfig{file=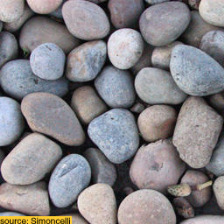, height = 4cm}\\
        \label{fig:texture_synthesis_a}
    \end{subfigure}%
    ~
    \begin{subfigure}[t]{0.33\textwidth}
        \centering
        \epsfig{file=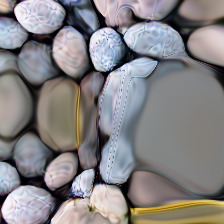, height = 4cm}\\
        \label{fig:texture_synthesis_b}
    \end{subfigure}%
    ~
    \begin{subfigure}[t]{0.33\textwidth}
        \centering
        \epsfig{file=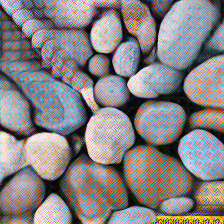, height = 4cm}\\
        \label{fig:texture_synthesis_c}
    \end{subfigure}%
    \caption{CPPNs vs.\ F-CPPNs for texture synthesis. (left) Target texture of pebbles. (middle) Output from \citet{mordvintsev2018differentiable}'s CPPN. (right) Output from our F-CPPN. By explicitly modelling frequencies beyond the DC component, our Fourier parameterization provides an improvement in surface detail on the synthesized pebbles.}
    \vspace{-3mm}
    \label{fig:texture_synthesis}
\end{figure*}

A visual texture can be loosely defined as a region of an image with stationary feature statistics. Examples of natural textures can include bark, granite, or sand. Texture synthesis is the process of algorithmically generating new image regions that match the stationary feature statistics of a given source texture. \citet{gatys2015texture} demonstrated impressive results using the learned filters from VGG-19. Textures were modelled in terms of the normalized correlations between activation maps within several layers of the network. Here we synthesize textures with our F-CPPN and the baseline CPPN using \citet{gatys2015texture}'s texture objective. The task is as follows. Given a target texture,
let $\mathbf{A}^{l} \in \mathbb{R}^{N_l\times M_l}$
be its row-vectorized activation maps at the $l$-th layer of a ConvNet (in this case, VGG-19). $N_l$ and $M_l$ denote the number of
activation maps and the number of spatial locations,
respectively.
The normalized correlations between activation maps
within a layer are encapsulated by a Gram matrix,
$\mathbf{G}^l \in \mathbb{R}^{N_l \times N_l}$, whose entries are given by,
$G_{ij}^l = \frac{1}{N_l M_l} \sum_{k=1}^{M_l} A_{ik}^l A_{jk}^l$.
$A_{ik}^l$ denotes the activation of feature $i$ at
location $k$ in layer $l$ on the target texture.
Similarly, given a synthesized texture, let $\hat{\mathbf{A}}^{l} \in \mathbb{R}^{N_l\times M_l}$ be its row-vectorized activation maps and $\hat{\mathbf{G}}^l \in \mathbb{R}^{N_l \times N_l}$ be its Gram matrix, whose entries are given by,
$\hat{G}_{ij}^l = \frac{1}{N_l M_l} \sum_{k=1}^{M_l} \hat{A}_{ik}^l \hat{A}_{jk}^l$.
The final objective is defined as the average of the MSE between
the Gram matrices of the target texture and that of the synthesized texture,
\begin{equation}
   \mathcal{L}_{style} = \frac{1}{L} \sum_{l} \Vert \mathbf{G}^l - \hat{\mathbf{G}}^l \Vert^2_F\ ,
   \label{eq:tex_loss}
\end{equation}
where $L$ is the number of ConvNet layers used when computing Gram matrices
and $\Vert \cdot \Vert_F$ is the Frobenius norm. This texture objective is also known as the style loss \cite{gatys2016image}. Similarly to \citet{gatys2015texture}, Gram matrices were computed on
layers \emph{conv1\_1}, \emph{pool1}, \emph{pool2}, \emph{pool3}, and \emph{pool4} of VGG-19.

In the case of \citet{gatys2015texture}, textures were synthesized by directly optimizing their pixel values. Recent approaches use generative ConvNets to synthesize textures~\cite{johnson2016perceptual,ulyanov2016texture}, parameterizing the output by the ConvNet's weights. Our implementation follows a similar approach, however, we use a F-CPPN as the generative network. This is a novel application of CPPNs.  Results are shown in Fig.~\ref{fig:texture_synthesis}. By explicitly modelling frequencies beyond the DC component, our F-CPPN provides an improvement in surface detail on the synthesized texture of pebbles. However, we observe a periodic tiling in the top-left region of the output. Whereas regions of constant output would correspond to untextured regions of constant colour in a CPPN (visible in CPPN output in Fig.~\ref{fig:texture_synthesis}), in a F-CPPN it would correspond to regions of the same coefficients, resulting in a $H_F \times W_F$-periodic tiling throughout the region.

\section{Conclusion}
\label{sec:conclusion}

In this paper, we presented an extension to CPPNs, based on Fourier analysis, which we call Fourier-CPPNs (F-CPPNs). F-CPPNs explicitly model the frequency information for each pixel output, capturing high-frequency detail that can not be captured by CPPNs. We applied our F-CPPN to the tasks of image reconstruction and texture synthesis and showed that the resulting images exhibited greater detail than the images synthesized by a CPPN. We observed a limitation common to both F-CPPNs and CPPNs where regions of constant output manifested themselves as regions of periodic tiling of texture in a F-CPPN's output and untextured regions of constant colour in a CPPN's output. Regularization methods may alleviate these issues, which we leave as directions for future work. An advantage of F-CPPNs is the direct manipulation of frequencies, allowing for interesting effects such as phase shifting \cite{freeman1991motion} and band-pass filtering. We aim to explore these techniques in future work.

{\small
\bibliographystyle{ieee}
\bibliography{main}

\begin{thebibliography}{17}
\providecommand{\natexlab}[1]{#1}
\providecommand{\url}[1]{\texttt{#1}}
\expandafter\ifx\csname urlstyle\endcsname\relax
  \providecommand{\doi}[1]{doi: #1}\else
  \providecommand{\doi}{doi: \begingroup \urlstyle{rm}\Url}\fi

\bibitem[Byrd et~al.(1995)Byrd, Lu, Nocedal, and Zhu]{byrd1995limited}
R.~H. Byrd, P.~Lu, J.~Nocedal, and C.~Zhu.
\newblock A limited memory algorithm for bound constrained optimization.
\newblock \emph{SIAM Journal on Scientific Computing}, 16\penalty0
  (5):\penalty0 1190--1208, 1995.

\bibitem[Cimpoi et~al.(2014)Cimpoi, Maji, Kokkinos, Mohamed, and
  Vedaldi]{cimpoi14describing}
M.~Cimpoi, S.~Maji, I.~Kokkinos, S.~Mohamed, and A.~Vedaldi.
\newblock Describing textures in the wild.
\newblock 2014.

\bibitem[Freeman et~al.(1991)Freeman, Adelson, and Heeger]{freeman1991motion}
W.~T. Freeman, E.~H. Adelson, and D.~J. Heeger.
\newblock Motion without movement.
\newblock 1991.

\bibitem[Gatys et~al.(2015)Gatys, Ecker, and Bethge]{gatys2015texture}
L.~A. Gatys, A.~S. Ecker, and M.~Bethge.
\newblock Texture synthesis using convolutional neural networks.
\newblock 2015.

\bibitem[Gatys et~al.(2016)Gatys, Ecker, and Bethge]{gatys2016image}
L.~A. Gatys, A.~S. Ecker, and M.~Bethge.
\newblock Image style transfer using convolutional neural networks.
\newblock 2016.

\bibitem[Goodfellow et~al.(2014)Goodfellow, Pouget-Abadie, Mirza, Xu,
  Warde-Farley, Ozair, Courville, and Bengio]{goodfellow2014generative}
I.~Goodfellow, J.~Pouget-Abadie, M.~Mirza, B.~Xu, D.~Warde-Farley, S.~Ozair,
  A.~Courville, and Y.~Bengio.
\newblock Generative adversarial nets.
\newblock 2014.

\bibitem[Ha(2016)]{ha2016generating}
D.~Ha.
\newblock Generating large images from latent vectors, 2016.

\bibitem[Johnson et~al.(2016)Johnson, Alahi, and
  Fei{-}Fei]{johnson2016perceptual}
J.~Johnson, A.~Alahi, and L.~Fei{-}Fei.
\newblock Perceptual losses for real-time style transfer and super-resolution.
\newblock 2016.

\bibitem[Karpathy(2014)]{karpathy2014image}
A.~Karpathy.
\newblock Image regression, 2014.

\bibitem[Karras et~al.(2018{\natexlab{a}})Karras, Aila, Laine, and
  Lehtinen]{karras2018progressive}
T.~Karras, T.~Aila, S.~Laine, and J.~Lehtinen.
\newblock Progressive growing of {GAN}s for improved quality, stability, and
  variation.
\newblock 2018{\natexlab{a}}.

\bibitem[Karras et~al.(2018{\natexlab{b}})Karras, Laine, and
  Aila]{karras2018style}
T.~Karras, S.~Laine, and T.~Aila.
\newblock A style-based generator architecture for generative adversarial
  networks.
\newblock \emph{arXiv}, 2018{\natexlab{b}}.

\bibitem[Metz and Gulrajani(2017)]{metz2017compositional}
L.~Metz and I.~Gulrajani.
\newblock Compositional pattern producing {GAN}.
\newblock 2017.

\bibitem[Mordvintsev et~al.(2018)Mordvintsev, Pezzotti, Schubert, and
  Olah]{mordvintsev2018differentiable}
A.~Mordvintsev, N.~Pezzotti, L.~Schubert, and C.~Olah.
\newblock Differentiable image parameterizations.
\newblock \emph{Distill}, 2018.

\bibitem[Simonyan and Zisserman(2014)]{simonyan2014very}
K.~Simonyan and A.~Zisserman.
\newblock Very deep convolutional networks for large-scale image recognition.
\newblock \emph{arXiv:1409.1556}, 2014.

\bibitem[Snelgrove and Tesfaldet(2018)]{snelgrove2018interactive}
X.~Snelgrove and M.~Tesfaldet.
\newblock Interactive {CPPN}s in {GLSL}.
\newblock 2018.

\bibitem[Stanley(2007)]{stanley2007compositional}
K.~O. Stanley.
\newblock Compositional pattern producing networks: A novel abstraction of
  development.
\newblock \emph{GPEV}, 8\penalty0 (2):\penalty0 131--162, 2007.

\bibitem[Ulyanov et~al.(2016)Ulyanov, Lebedev, Vedaldi, and
  Lempitsky]{ulyanov2016texture}
D.~Ulyanov, V.~Lebedev, A.~Vedaldi, and V.~S. Lempitsky.
\newblock Texture networks: {F}eed-forward synthesis of textures and stylized
  images.
\newblock 2016.

\end{thebibliography}
}

\appendix

\section{Additional results for image reconstruction}
See Tables \ref{tab:image_reconstruction_sup_a}, \ref{tab:image_reconstruction_sup_b}, \ref{tab:image_reconstruction_sup_c}, and \ref{tab:image_reconstruction_sup_d} for further image reconstruction results with a CPPN and our F-CPPN on images from the Describable Textures Dataset (DTD) \cite{cimpoi14describing}. Recall that the optimized objective is the L2 distance between activations of intermediate layers of a pre-trained VGG-19 network \cite{simonyan2014very}.

Note the results for the honeycomb image in the first row of Table \ref{tab:image_reconstruction_sup_a}. The CPPN appears
to have insufficient capacity to resolve all of the high-frequency
edges of the honeycomb lattice, and so there is a patch that it fails
to reconstruct. Our F-CPPN appears to not have the same capacity limitations. We see similar regions where the CPPN fails to reconstruct
a portion of the edges of a target image in the second and third rows as well. Note also the ``smeared'' quality in many of the CPPN images such as the paisley image on the fourth row of Table \ref{tab:image_reconstruction_sup_a}, the scaly image in the second row of Table \ref{tab:image_reconstruction_sup_d}, and the tiger fur image in the third row of Table \ref{tab:image_reconstruction_sup_d}. In all cases, our F-CPPN has a lower MSE (average MSE of $1.07e5$ compared to CPPN's average MSE of $1.6e5$), showing that it is not only qualitatively better, but also performs quantitatively better on this particular objective.

\section{Additional results for texture synthesis}
Both the CPPN and F-CPPN have more trouble with the texture synthesis objective than with the image reconstruction objective, as shown in Tables \ref{tab:texture_synthesis_sup_a} and \ref{tab:texture_synthesis_sup_b}. Images from the DTD \cite{cimpoi14describing} are used as target textures.

While the CPPN tends to create large regions of constant or smoothly varying colour, the F-CPPN is able to fill these regions with periodic texture containing some of the qualities of the original image. Its results are consistently qualitatively superior than the CPPN's results. For example, for the bubbles texture in the second row of Table \ref{tab:texture_synthesis_sup_b}, the CPPN captures very few qualities of the original image beyond its approximate colour scheme, whereas the F-CPPN reconstructs both the small and large bubbles, as well as the bubble-like periodic texture.

Although our F-CPPN is not yet achieving the perceptual quality of state-of-the-art texture synthesis algorithms \cite{gatys2015texture,ulyanov2016texture}, it appears both CPPNs and F-CPPNs are failing to escape local optima. We note that F-CPPNs are able to impressively reconstruct images, down to their higher-frequency textural detail. This leads us to believe that by modifying the texture synthesis objective to take better advantage of the architecture of F-CPPNs, we can improve optimization results. We leave this for future work.

\section{Latent space interpolation}
Recall that CPPNs and F-CPPNs accept $(x, y)$ pixel coordinates as input. Up until this point, we have optimized our F-CPPN on an objective with a single target. We briefly experimented with optimizing our F-CPPN on multiple targets. This was achieved by modifying the $(x, y)$ coordinate input by including an additional variable $\vec{z}$ that conditions the output of the F-CPPN. $z$ is constant for all $(x, y)$ inputs. We optimized our F-CPPN with the texture synthesis objective against two target textures. The two targets were an image of pebbles and an image of peppers, with $\vec{z}=(1,0)$ and $\vec{z}=(0,1)$, respectively.

\begin{figure*}[t]
    \centering
    \begin{subfigure}[t]{0.25\textwidth}
        \centering
        \epsfig{file=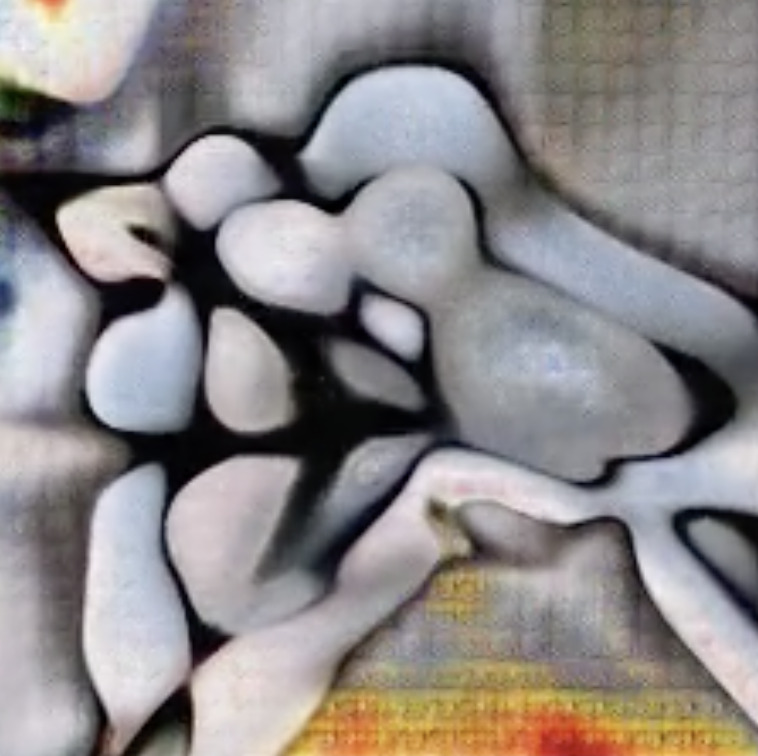, height = 4cm}\\
    \end{subfigure}%
    ~
    \begin{subfigure}[t]{0.25\textwidth}
        \centering
        \epsfig{file=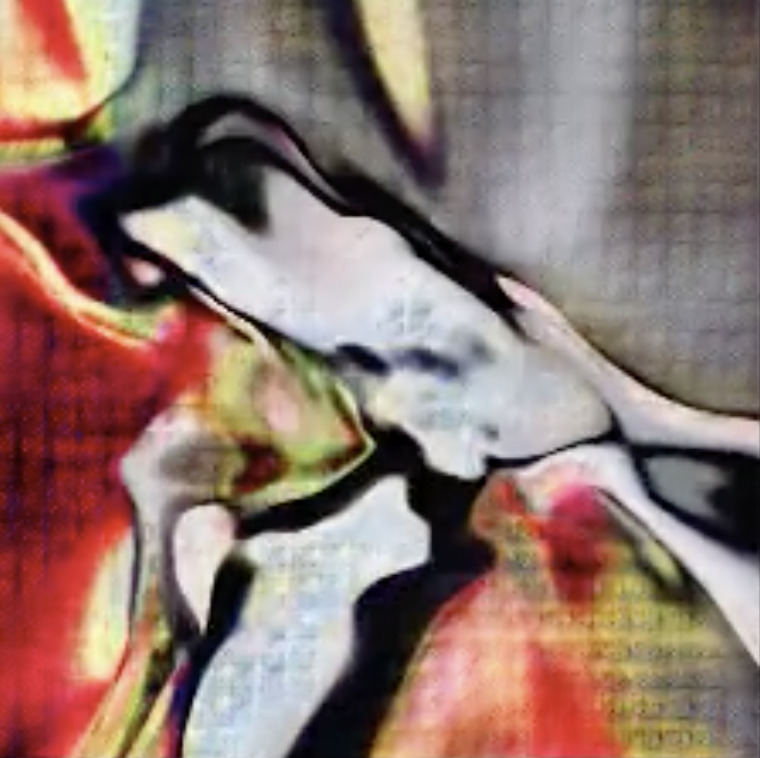, height = 4cm}\\
    \end{subfigure}%
    ~
    \begin{subfigure}[t]{0.25\textwidth}
        \centering
        \epsfig{file=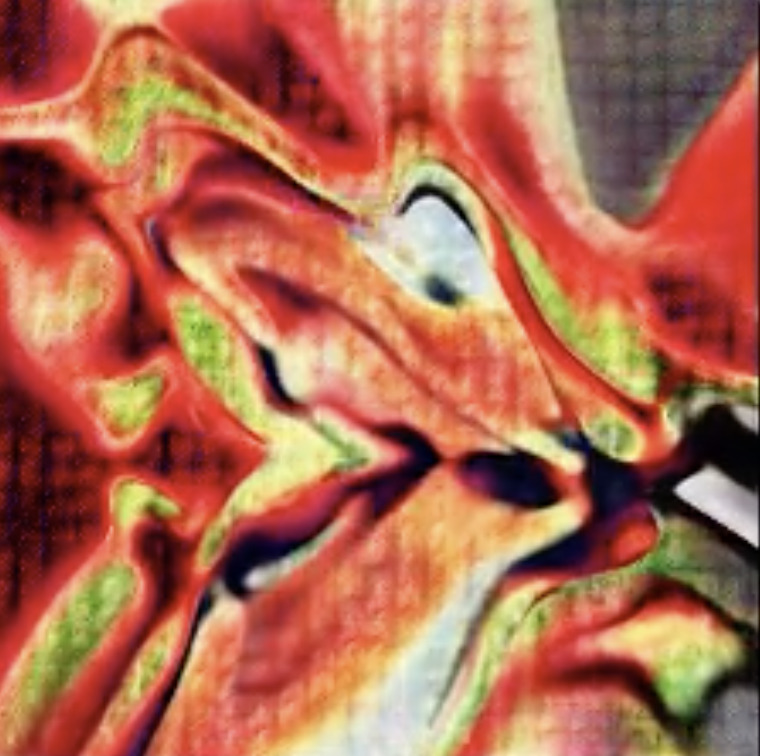, height = 4cm}\\
    \end{subfigure}%
    ~
    \begin{subfigure}[t]{0.25\textwidth}
        \centering
        \epsfig{file=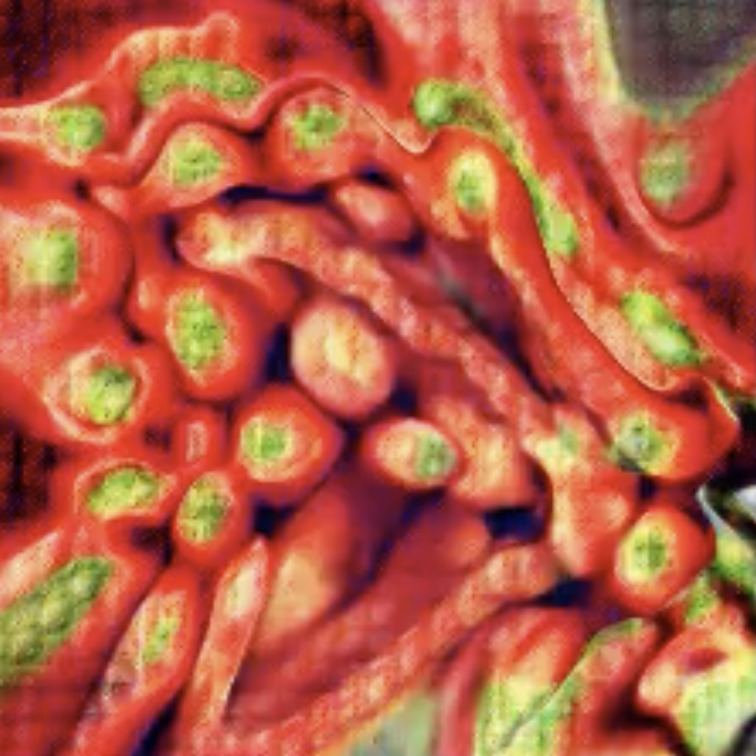, height = 4cm}\\
    \end{subfigure}%
    \caption{F-CPPN latent space interpolation. Instead of a $(x,y)$ pixel coordinate input, the F-CPPN accepts a $(x,y,\vec{z})$ input where $\vec{z}$ is a variable that conditions the output of the F-CPPN based on a specific target. The two targets in this case are an image of pebbles ($\vec{z} = (1, 0)$) and an image of peppers ($\vec{z} = (0, 1)$). (left-to-right) conditioned output as $\vec{z}$ is interpolated with the function $\vec{z}=(\cos{\theta}, \sin{\theta})$ and $\theta$ is incremented from $0$ to $\pi/2$. The F-CPPN was optimized with the texture synthesis objective.}
    \label{fig:latent_space_interpolation}
\end{figure*}

Figure \ref{fig:latent_space_interpolation} shows our conditioned F-CPPN's output during inference time, where $\vec{z}$ is interpolated for each frame with the function $\vec{z} = (\cos{\theta}, \sin{\theta})$ and $\theta$ is incremented from $0$ to $\pi/2$. The result is an aesthetically pleasing video of a texture of pebbles warping to a texture of peppers.

\section{High-resolution synthesis}
\begin{figure*}[t]
    \centering
    \begin{subfigure}[t]{0.5\textwidth}
        \centering
        \epsfig{file=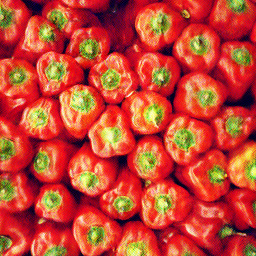, height = 4cm}\\
        \label{fig:high_resolution_image_reconstruction_a}
    \end{subfigure}%
    ~
    \begin{subfigure}[t]{0.5\textwidth}
        \centering
        \epsfig{file=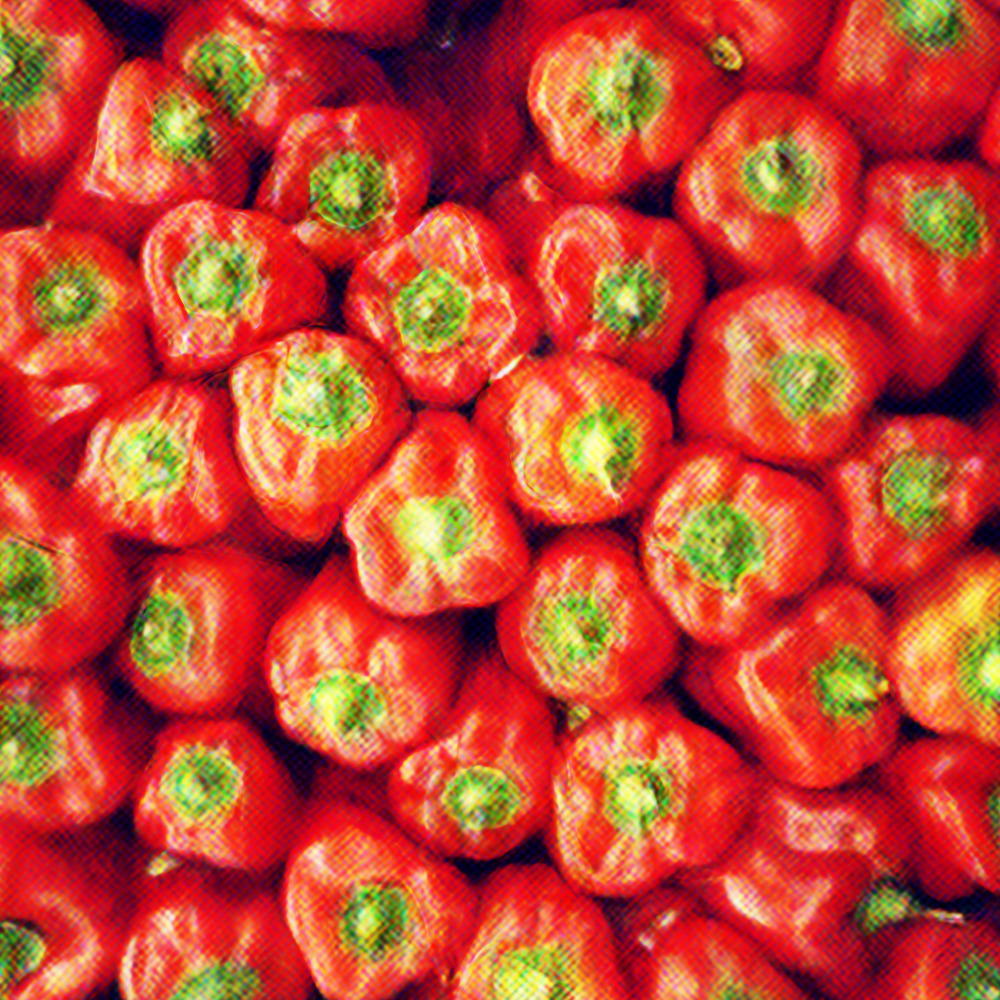, height = 8cm}\\
        \label{fig:high_resolution_image_reconstruction_b}
    \end{subfigure}%
    \caption{High-resolution image reconstruction using F-CPPNs. (left) A $256 \times 256$ image synthesized by a F-CPPN. (right) A $1000 \times 1000$ image synthesized by a F-CPPN. CPPNs and F-CPPNs have the benefit of being able to directly synthesize images at any resolution, including resolutions other than the ones they were optimized at. The F-CPPN was optimized with the image reconstruction objective and the resolution of the target was $256 \times 256$.}
    \vspace{-3mm}
    \label{fig:high_resolution_image_reconstruction}
\end{figure*}

\begin{figure*}[t]
    \centering
    \begin{subfigure}[t]{0.5\textwidth}
        \centering
        \epsfig{file=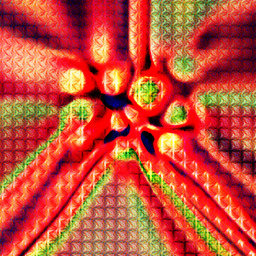, height = 4cm}\\
        \label{fig:high_resolution_texture_synthesis_a}
    \end{subfigure}%
    ~
    \begin{subfigure}[t]{0.5\textwidth}
        \centering
        \epsfig{file=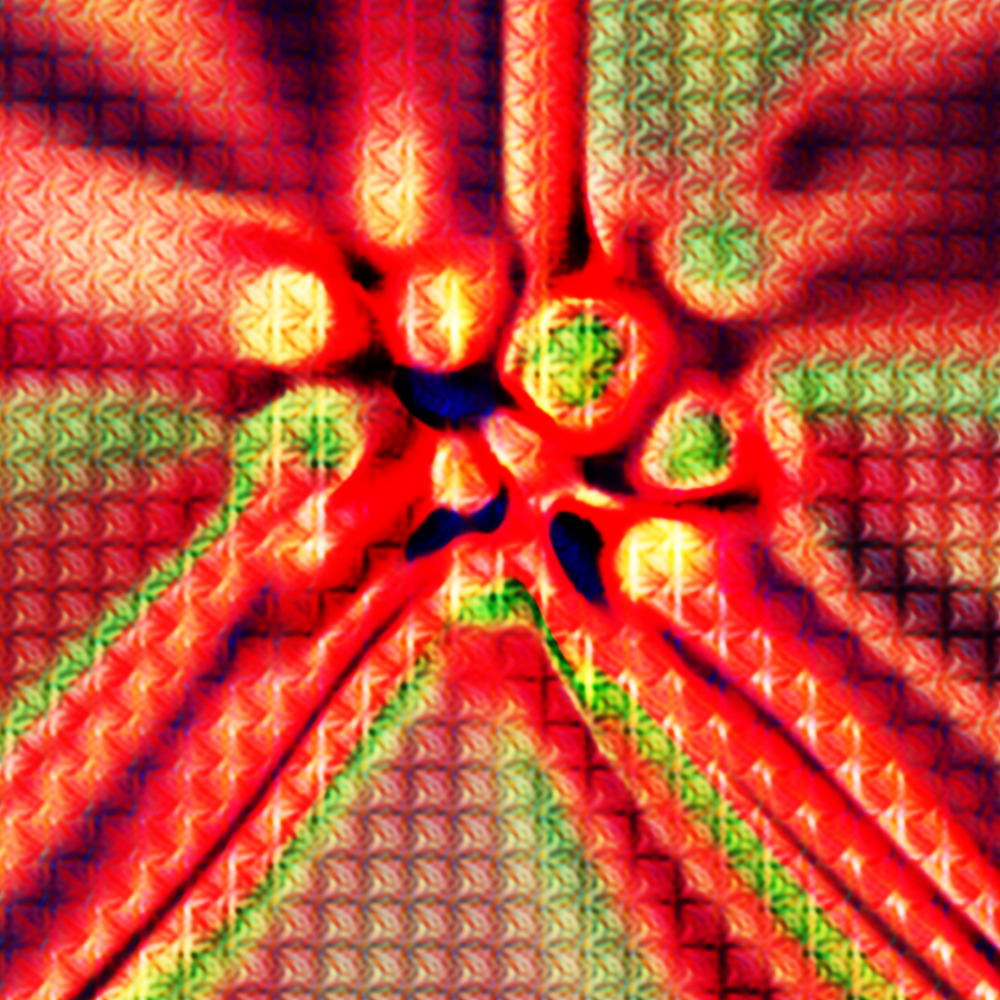, height = 8cm}\\
        \label{fig:high_resolution_texture_synthesis_b}
    \end{subfigure}%
    \caption{High-resolution texture synthesis using F-CPPNs. (left) A $256 \times 256$ texture synthesized by a F-CPPN. (right) A $1000 \times 1000$ texture synthesized by a F-CPPN. CPPNs and F-CPPNs have the benefit of being able to directly synthesize images at any resolution, including resolutions other than the ones they were optimized at. The F-CPPN was optimized with the texture synthesis objective and the resolution of the target was $256 \times 256$.}
    \vspace{-3mm}
    \label{fig:high_resolution_texture_synthesis}
\end{figure*}

CPPNs and F-CPPNs are continuous functions that map $(x,y)$ pixel coordinates to $(r,g,b)$ colours; they are not inherently tied to the pixel grid. By feeding sub-pixel coordinates, these continuous functions naturally interpolate pixel colours. This means that high-resolution zoomed-in images can be generated by simply feeding in a more densely sampled coordinate grid as the input. An example of this is shown in Fig.~\ref{fig:high_resolution_image_reconstruction} and \ref{fig:high_resolution_texture_synthesis}, for both the image reconstruction and texture synthesis objective, respectively.

Note however, that the highest spatial frequency the F\nobreakdash-CPPN can output is a hyperparameter of the network architecture itself, so no higher-frequency details other than those visible at the original scale are synthesized as the image is super-sampled. Eventually, the image will appear smooth as the zoom level is increased. This is analogous to the ``infinite-resolution'' of vector graphics, where the image does not become ``pixelated'' as the zoom level increases, but no new details are introduced.

\begin{table*}[t]
    \centering
    \begin{tabular}{|c|c|c|}
        \hline
        Target image & CPPN & F-CPPN (ours) \\ [0.5ex]
        \hline\hline
        &  &  \\
        \epsfig{file=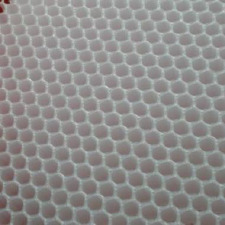, height = 4cm} & \epsfig{file=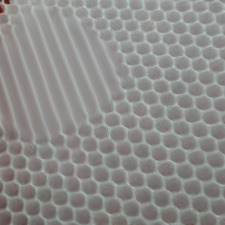, height = 4cm} & \epsfig{file=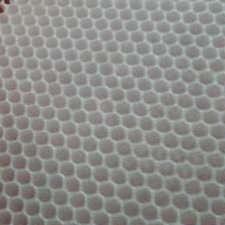, height = 4cm} \\
        & $1.71e4$ & $\mathbf{0.81e4}$ \\
        \hline
        &  &  \\
        \epsfig{file=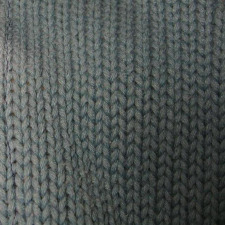, height = 4cm} & \epsfig{file=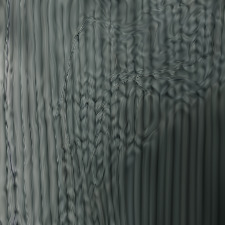, height = 4cm} & \epsfig{file=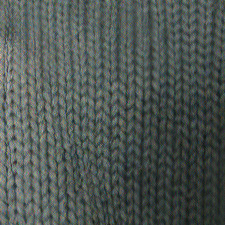, height = 4cm} \\
        & $5.77e4$ & $\mathbf{3.25e4}$ \\
        \hline
        &  &  \\
        \epsfig{file=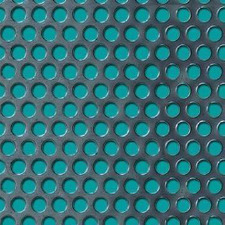, height = 4cm} & \epsfig{file=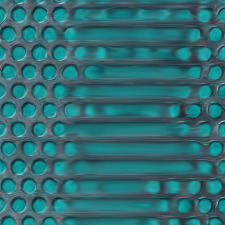, height = 4cm} & \epsfig{file=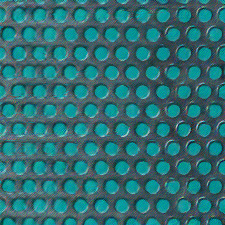, height = 4cm} \\
        & $1.48e5$ & $\mathbf{4.98e4}$ \\
        \hline
        &  &  \\
        \epsfig{file=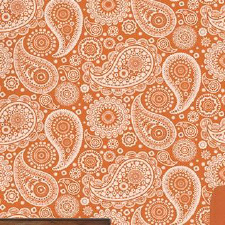, height = 4cm} & \epsfig{file=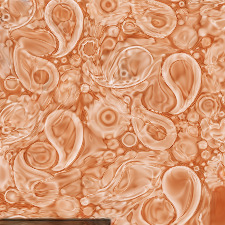, height = 4cm} & \epsfig{file=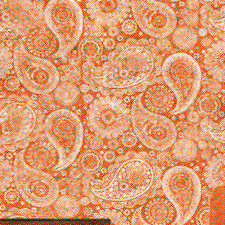, height = 4cm} \\
        & $1.84e5$ & $\mathbf{1.47e5}$ \\
        \hline
    \end{tabular}
    \caption{Image reconstruction via Compositional Pattern Producing Networks (CPPNs) and Fourier-CPPNs (F-CPPNs). (left) Target image. (middle) CPPN output using the same CPPN architecture as \citet{mordvintsev2018differentiable}. (right) Our F-CPPN's output. Numbers beneath images indicate image reconstruction error (\ie, MSE).}
    \label{tab:image_reconstruction_sup_a}
\end{table*}

\begin{table*}[t]
    \centering
    \begin{tabular}{|c|c|c|}
        \hline
        Target image & CPPN & F-CPPN (ours) \\ [0.5ex]
        \hline\hline
        &  &  \\
        \epsfig{file=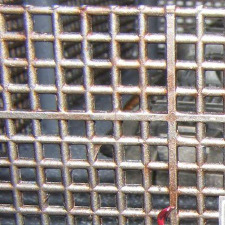, height = 4cm} & \epsfig{file=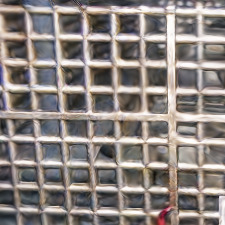, height = 4cm} & \epsfig{file=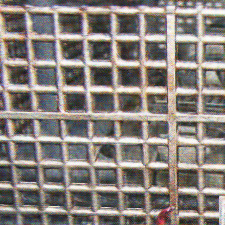, height = 4cm} \\
        & $7.28e4$ & $\mathbf{5.83e4}$ \\
        \hline
        &  &  \\
        \epsfig{file=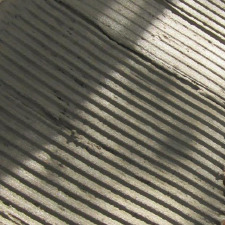, height = 4cm} & \epsfig{file=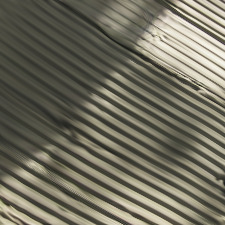, height = 4cm} & \epsfig{file=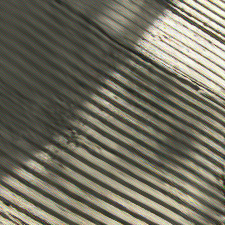, height = 4cm} \\
        & $3.31e4$ & $\mathbf{2.79e4}$ \\
        \hline
        &  &  \\
        \epsfig{file=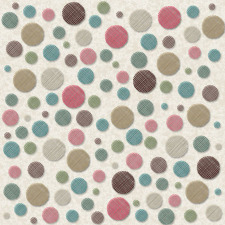, height = 4cm} & \epsfig{file=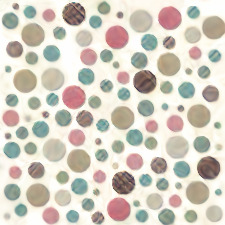, height = 4cm} & \epsfig{file=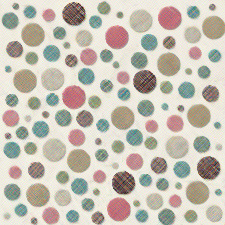, height = 4cm} \\
        & $3.8e4$ & $\mathbf{2.26e4}$ \\
        \hline
        &  &  \\
       \epsfig{file=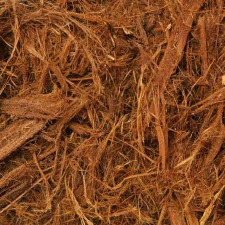, height = 4cm} & \epsfig{file=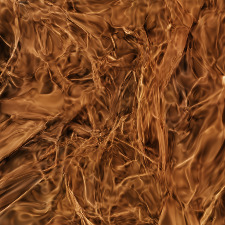, height = 4cm} & \epsfig{file=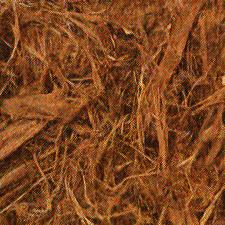, height = 4cm} \\
        & $9.83e4$ & $\mathbf{8.76e4}$ \\
        \hline
    \end{tabular}
    \caption{Image reconstruction via Compositional Pattern Producing Networks (CPPNs) and Fourier-CPPNs (F-CPPNs). (left) Target image. (middle) CPPN output using the same CPPN architecture as \citet{mordvintsev2018differentiable}. (right) Our F-CPPN's output. Numbers beneath images indicate image reconstruction error (\ie, MSE).}
    \label{tab:image_reconstruction_sup_b}
\end{table*}

\begin{table*}[t]
    \centering
    \begin{tabular}{|c|c|c|}
        \hline
        Target image & CPPN & F-CPPN (ours) \\ [0.5ex]
        \hline\hline
        &  &  \\
        \epsfig{file=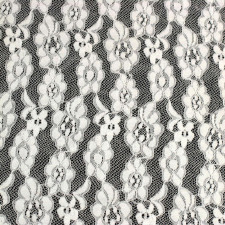, height = 4cm} & \epsfig{file=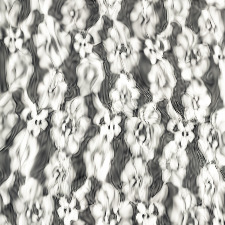, height = 4cm} & \epsfig{file=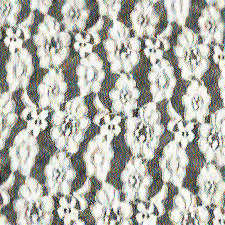, height = 4cm} \\
        & $1.99e5$ & $\mathbf{1.52e5}$ \\
        \hline
        &  &  \\
        \epsfig{file=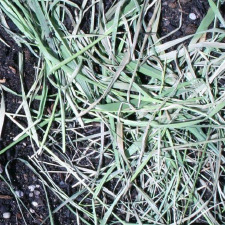, height = 4cm} & \epsfig{file=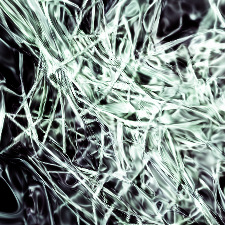, height = 4cm} & \epsfig{file=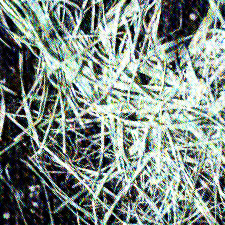, height = 4cm} \\
        & $4.31e5$ & $\mathbf{3.39e5}$ \\
        \hline
        &  &  \\
        \epsfig{file=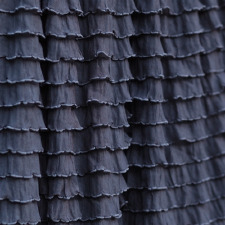, height = 4cm} & \epsfig{file=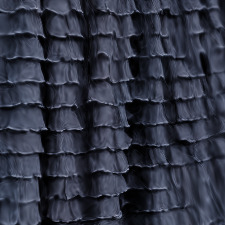, height = 4cm} & \epsfig{file=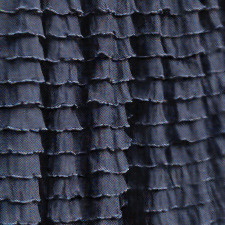, height = 4cm} \\
        & $3.19e4$ & $\mathbf{2.24e4}$ \\
        \hline
        &  &  \\
        \epsfig{file=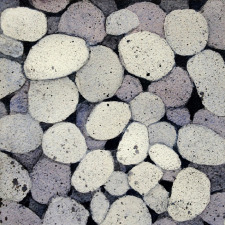, height = 4cm} & \epsfig{file=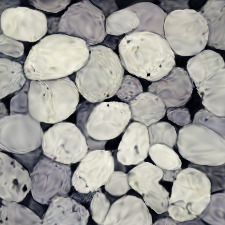, height = 4cm} & \epsfig{file=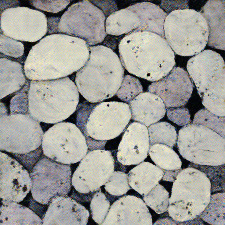, height = 4cm} \\
        & $7.55e4$ & $\mathbf{5.76e4}$ \\
        \hline
    \end{tabular}
    \caption{Image reconstruction via Compositional Pattern Producing Networks (CPPNs) and Fourier-CPPNs (F-CPPNs). (left) Target image. (middle) CPPN output using the same CPPN architecture as \citet{mordvintsev2018differentiable}. (right) Our F-CPPN's output. Numbers beneath images indicate image reconstruction error (\ie, MSE).}
    \label{tab:image_reconstruction_sup_c}
\end{table*}

\begin{table*}[t]
    \centering
    \begin{tabular}{|c|c|c|}
        \hline
        Target image & CPPN & F-CPPN (ours) \\ [0.5ex]
        \hline\hline
        &  &  \\
        \epsfig{file=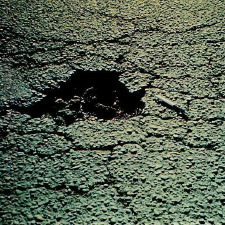, height = 4cm} & \epsfig{file=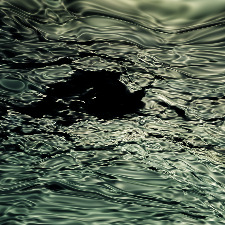, height = 4cm} & \epsfig{file=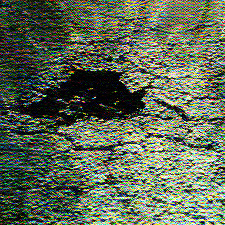, height = 4cm} \\
        & $7.41e5$ & $\mathbf{4.8e5}$ \\
        \hline
        &  &  \\
        \epsfig{file=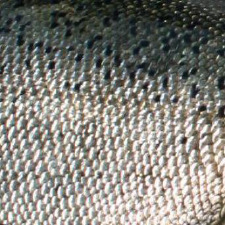, height = 4cm} & \epsfig{file=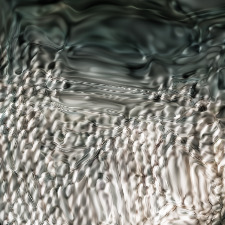, height = 4cm} & \epsfig{file=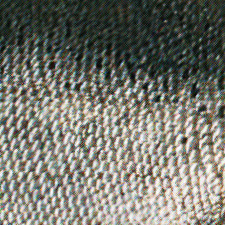, height = 4cm} \\
        & $2.1e5$ & $\mathbf{8.16e4}$ \\
        \hline
        &  &  \\
        \epsfig{file=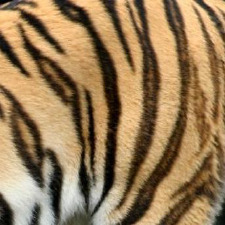, height = 4cm} & \epsfig{file=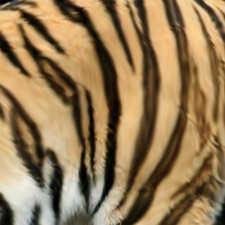, height = 4cm} & \epsfig{file=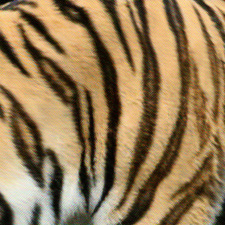, height = 4cm} \\
        & $3.36e4$ & $\mathbf{1.82e4}$ \\
        \hline
        &  &  \\
        \epsfig{file=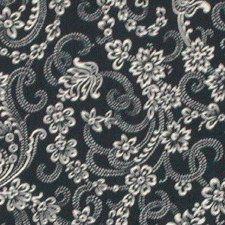, height = 4cm} & \epsfig{file=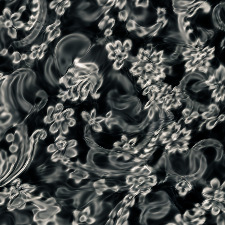, height = 4cm} & \epsfig{file=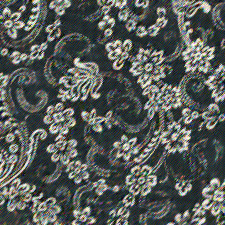, height = 4cm} \\
        & $1.89e5$ & $\mathbf{1.28e5}$ \\
        \hline
    \end{tabular}
    \caption{Image reconstruction via Compositional Pattern Producing Networks (CPPNs) and Fourier-CPPNs (F-CPPNs). (left) Target image. (middle) CPPN output using the same CPPN architecture as \citet{mordvintsev2018differentiable}. (right) Our F-CPPN's output. Numbers beneath images indicate image reconstruction error (\ie, MSE).}
    \label{tab:image_reconstruction_sup_d}
\end{table*}
\begin{table*}[t]
    \centering
    \begin{tabular}{|c|c|c|}
        \hline
        Target texture & CPPN & F-CPPN (ours) \\ [0.5ex]
        \hline\hline
        &  &  \\
        \epsfig{file=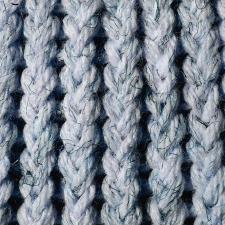, height = 4cm} & \epsfig{file=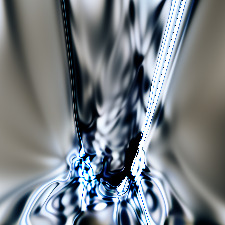, height = 4cm} & \epsfig{file=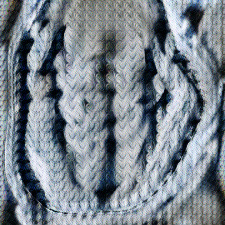, height = 4cm} \\
        &  &  \\
        \hline
        &  &  \\
        \epsfig{file=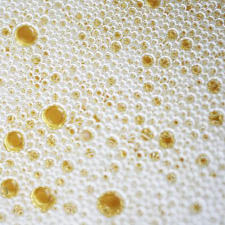, height = 4cm} & \epsfig{file=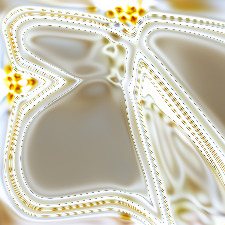, height = 4cm} & \epsfig{file=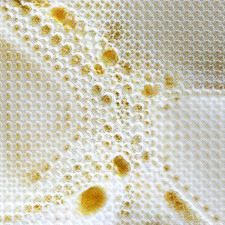, height = 4cm} \\
        &  &  \\
        \hline
        &  &  \\
        \epsfig{file=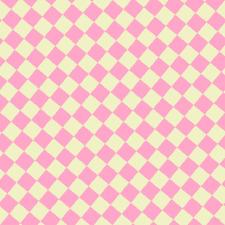, height = 4cm} & \epsfig{file=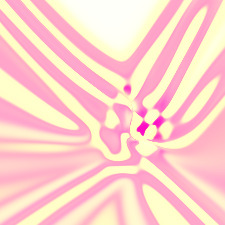, height = 4cm} & \epsfig{file=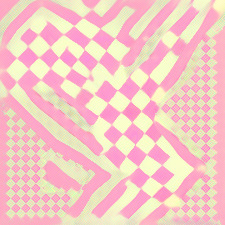, height = 4cm} \\
        &  &  \\
        \hline
        &  &  \\
        \epsfig{file=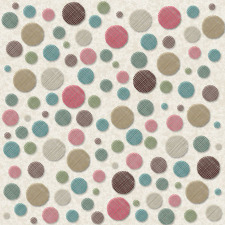, height = 4cm} & \epsfig{file=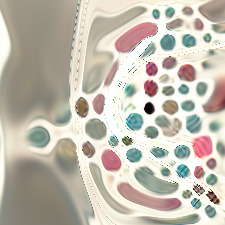, height = 4cm} & \epsfig{file=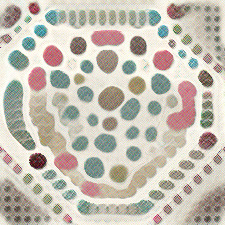, height = 4cm} \\
        &  &  \\
        \hline
    \end{tabular}
    \caption{Texture synthesis via Compositional Pattern Producing Networks (CPPNs) and Fourier-CPPNs (F-CPPNs). (left) Target texture. (middle) CPPN output using the same CPPN architecture as \citet{mordvintsev2018differentiable}. (right) Our F-CPPN's output.}
    \label{tab:texture_synthesis_sup_a}
\end{table*}

\begin{table*}[t]
    \centering
    \begin{tabular}{|c|c|c|}
        \hline
        Target texture & CPPN & F-CPPN (ours) \\ [0.5ex]
        \hline\hline
        &  &  \\
        \epsfig{file=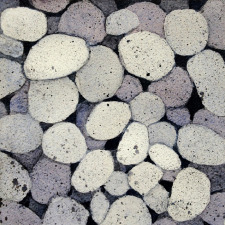, height = 4cm} & \epsfig{file=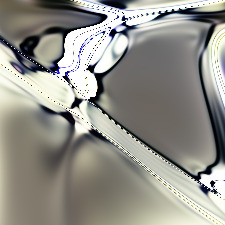, height = 4cm} & \epsfig{file=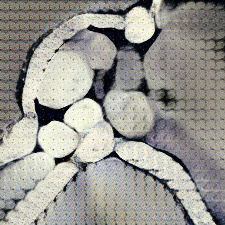, height = 4cm} \\
        &  &  \\
        \hline
        &  &  \\
        \epsfig{file=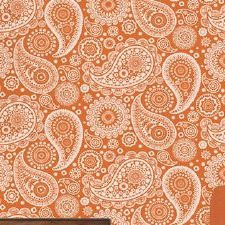, height = 4cm} & \epsfig{file=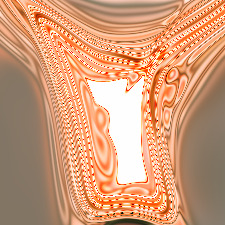, height = 4cm} & \epsfig{file=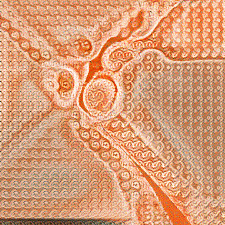, height = 4cm} \\
        &  &  \\
        \hline
        &  &  \\
        \epsfig{file=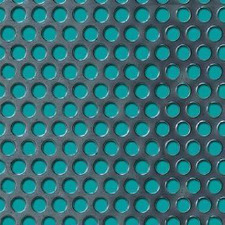, height = 4cm} & \epsfig{file=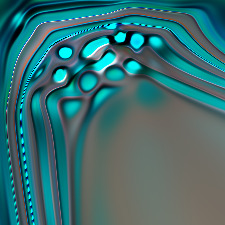, height = 4cm} & \epsfig{file=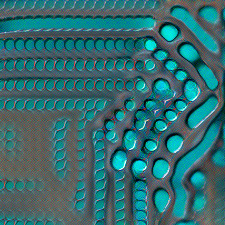, height = 4cm} \\
        &  &  \\
        \hline
        &  &  \\
        \epsfig{file=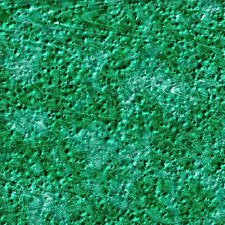, height = 4cm} & \epsfig{file=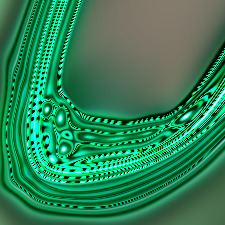, height = 4cm} & \epsfig{file=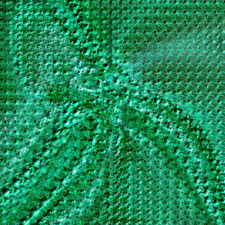, height = 4cm} \\
        &  &  \\
        \hline
    \end{tabular}
    \caption{Texture synthesis via Compositional Pattern Producing Networks (CPPNs) and Fourier-CPPNs (F-CPPNs). (left) Target texture. (middle) CPPN output using the same CPPN architecture as \citet{mordvintsev2018differentiable}. (right) Our F-CPPN's output.}
    \label{tab:texture_synthesis_sup_b}
\end{table*}

\end{document}